\documentclass[10pt,twocolumn,letterpaper]{article}

\usepackage[pagenumbers]{cvpr} 

\definecolor{cvprblue}{rgb}{0.21,0.49,0.74}
\usepackage[pagebackref,breaklinks,colorlinks,allcolors=cvprblue]{hyperref}
\usepackage{multirow}
\usepackage{color, colortbl}
\usepackage{fontawesome5}
\usepackage{pifont}
\usepackage{float} 
\usepackage{pifont}
	
\definecolor{LightCyan}{rgb}{0.88,1,1}

\def\paperID{4887} 
\def\confName{CVPR}
\def\confYear{2025}

\newcommand*{\affmark}[1][*]{\textsuperscript{#1}}

\newcommand*{\affaddr}[1]{#1}

\title{EAGLE: Enhanced Visual Grounding Minimizes Hallucinations in Instructional Multimodal Models}

\newcommand{\methodname}{EAGLE }

\author{%
Andrés Villa\affmark[1], Juan León  Alcázar\affmark[1], Motasem Alfarra\affmark[1], Vladimir Araujo\affmark[2], \\Alvaro Soto\affmark[3], Bernard Ghanem\affmark[1]
\\
\normalsize \small\affaddr{\affmark[1]King Abdullah University of Science and Technology (KAUST)}, \\ \normalsize \small\affaddr{\affmark[2]Sailplane AI, \affmark[3]Pontificia Universidad Católica de Chile}
}

\begin{document}
\twocolumn[{
\renewcommand\twocolumn[1][]{#1}
\maketitle
\thispagestyle{empty}
\begin{center}
  \vspace{-0.6cm}
  \newcommand{\teaserwidth}{\textwidth}
  \centerline{
    \includegraphics[width=0.96\teaserwidth]{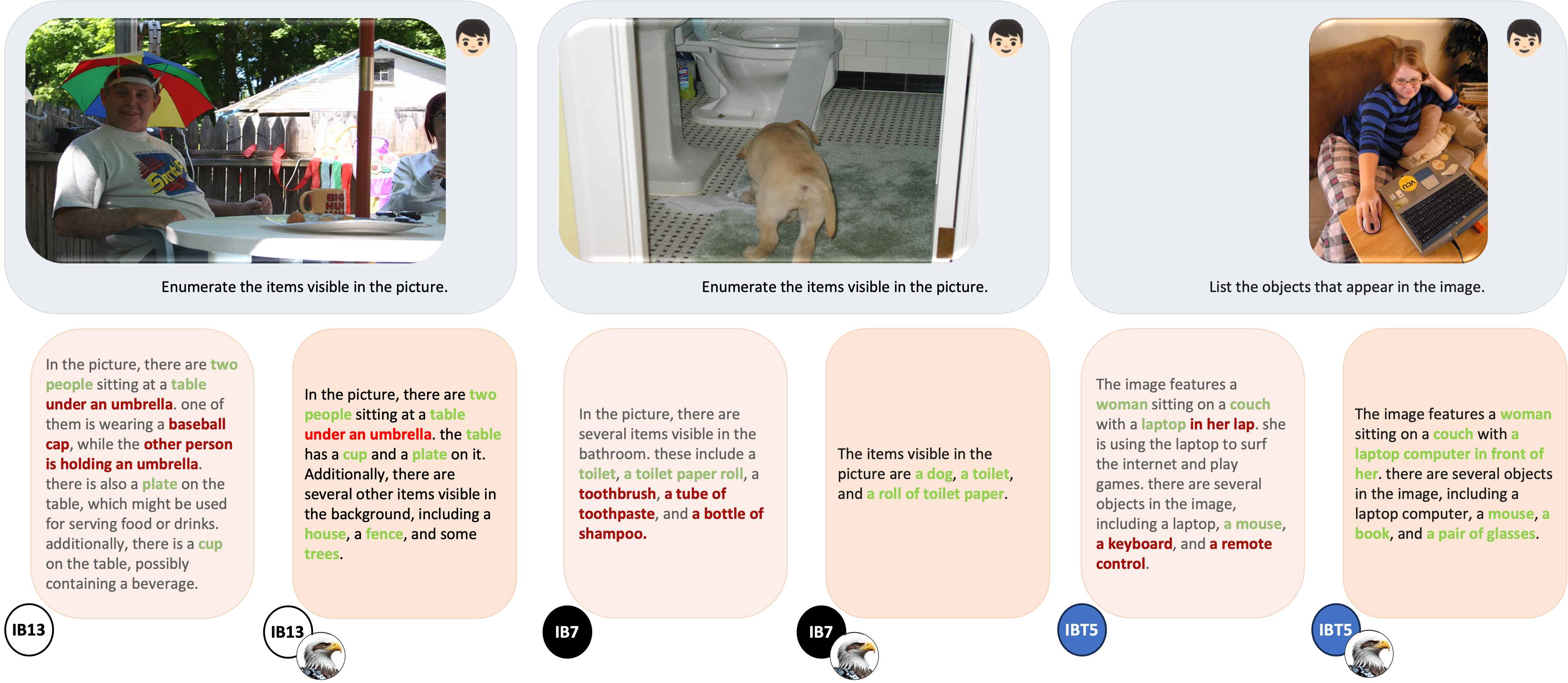}
    \vspace{-0.2cm}
    }
    \captionof{figure}{\textbf{\methodname visual encoders reduce hallucinations in IT-VLMs.} We present three example scenarios, each featuring a question about an image input to a specific IT-VLM with its original visual encoder (left  box in pink) and the corresponding EAGLE-tuned visual encoder (right box in orange). “IB7", “IB13" and “IBT5" refer to InstructBLIP with Vicuna7B, Vicuna13B, and FlanT5xl, respectively. EAGLE substantially reduces hallucinations, providing more visually grounded and reliable descriptions. \textbf{i)} In the first example (left), IB13 with the EAGLE-tuned visual encoder can clearly identify fine-grained elements such as a fence, house, and trees. \textbf{ii)} In the second example (center), EAGLE helps IB7 to accurately recognize a dog, even in an unusual viewpoint and context. \textbf{iii)} In the third example (right), EAGLE enhances object localization, allowing the model to precisely identify the laptop’s position. These examples illustrate EAGLE’s effectiveness in improving visual grounding across complex, multi-object scenes.}%
    \vspace{-0.1cm}
  \label{fig:teaser_fig}
 \end{center}
}]

\begin{abstract}

%
Large language models and vision transformers have demonstrated impressive zero-shot capabilities, enabling significant transferability in downstream tasks. The fusion of these models has resulted in multi-modal architectures with enhanced instructional capabilities. Despite incorporating vast image and language pre-training, these multi-modal architectures often generate responses that deviate from the ground truth in the image data. These failure cases are known as hallucinations. Current methods for mitigating hallucinations generally focus on regularizing the language component, improving the fusion module, or ensembling multiple visual encoders to improve visual representation. In this paper, we address the hallucination issue by directly enhancing the capabilities of the visual component. Our approach, named \methodname, is fully agnostic to the LLM or fusion module and works as a post-pretraining approach that improves the grounding and language alignment of the visual encoder. We show that a straightforward reformulation of the original contrastive pre-training task results in an improved visual encoder that can be incorporated into the instructional multi-modal architecture without additional instructional training. As a result, \methodname achieves a significant reduction in hallucinations across multiple challenging benchmarks and tasks.
\end{abstract}    
\section{Introduction}
\label{sec:intro}

Large-scale pre-trained architectures have significantly advanced the fields of Natural Language Processing (NLP), and Computer Vision. In the language domain, Large Language Models (LLMs) have demonstrated remarkable zero-shot performance in multiple NLP tasks  \cite{brown2020language, devlin2018bert, liu2019roberta, raffel2020exploring, vicuna2023, chung2024scaling, touvron2023llama}. This success has resulted in a shift in NLP research, moving from task-specific pre-training to task-agnostic representation learning. In the image domain, visual transformers (ViT) \cite{dosovitskiy2020image} have driven significant advances in zero-shot transferability \cite{pmlr-v139-radford21a, sun2023evaclipimprovedtrainingtechniques, oquab2023dinov2, zhai2023sigmoid, pmlr-v119-chen20j, grill2020bootstrap, He_2020_CVPR, cherti2023reproducible}, and enabled new state-of-the-art results on downstream tasks.

Building upon the improvements attained by transformer architectures in vision and NLP, Instruction Tuning Vision and Language Models (IT-VLM) have emerged as the ensemble of an LLM and a large-scale ViT fine-tuned with instructional data. IT-VLMs leverage the zero-shot capabilities of both modalities by learning a small fusion sub-network known as adapter or connector \cite{blip, liu2023improvedllava, dai2024instructblip, zhu2023minigpt, kosmos2,Qwen-VL, internlmxcomposer2, blip3}. Although the fusion module is far smaller than the other components, it enables IT-VLM models to achieve notable zero-shot performance in multi-modal tasks, such as VQA, Image Captioning, and Image Classification.

The open-set natural language responses of IT-VLM allow for the generation of erroneous outputs unrelated to image ground-truth, these false positive errors are known as \textit{hallucinations} \cite{villa2024magicmerlimmultimodalevaluation, li2023evaluating, Tong_2024_CVPR, guan2023hallusionbench}. This phenomenon is often a result of language biases in the decoder module, as the decoding component would generate outputs consistent with the current stream of generated tokens while ignoring the semantic information in the visual modality \cite{villa2024magicmerlimmultimodalevaluation}.

Existing methods for reducing hallucinations in IT-VLMs primarily focus on the language model (LLM) component or the adapter module. These methods include optimizing the adapter module’s training strategy \cite{kosmos2, Qwen-VL}, improving the adapter module's architecture \cite{blip, dai2024instructblip, liu2023improvedllava}, leveraging extensive data to pre-train robust LLMs \cite{chen2022pali, Qwen-VL, liu2023improvedllava}, and enhancing instructional training data through commercial LLMs \cite{liu2023improvedllava, liu2023visual}. In contrast, we address the hallucination problem from an orthogonal direction by focusing exclusively on the visual component of the IT-VLM. 

Previous approaches to enhance the visual representation in IT-VLMs \cite{Tong_2024_CVPR, kar2024brave} introduce additional visual encoders, which prevent scalability and require further fine-tuning of the IT-VLM. We depart from that practice and develop a method that directly enforces visual grounding on the Vision Transformer (ViT) for a subset of common object classes while preserving the global feature descriptor. This refined ViT effectively encodes fine-grained object class distributions and seamlessly integrates into an IT-VLM model without additional fine-tuning or adaptation. Empirical results confirm that our ViT module consistently reduces hallucinations across diverse benchmarks, demonstrating compatibility and effectiveness across IT-VLM architectures. In Figure \ref{fig:teaser_fig}, we provide some qualitative examples that show the baseline (left pink box) and improved response (right orange box) of IT-VLMs. Notably, when our ViT is incorporated, we observe significantly more factual and visually grounded responses.


In this paper, we introduce \textbf{E}nhanced Visu\textbf{A}l \textbf{G}rounding Minimizes Hallucinations in Instructional Mu\textbf{L}timodal Mod\textbf{E}ls (EAGLE), a pre-training strategy that significantly reduces the hallucinations in IT-VLMs. Unlike other methods, \methodname does not target a specific IT-VLM architecture, or a particular LLM. In fact, \methodname is fully agnostic to the other IT-VLM components. At training time, \methodname optimizes the ViT component in isolation from the LLM and the fusion module; the improved visual transformer is then plugged into the trained IT-VLM at inference time without any further modification. Without any bells and whistles, our tuned ViTs reduce the hallucinations in common IT-VLMs. Across three well-established hallucination benchmarks and 6 different IT-VLMs models, we show direct improvement by simply replacing their default ViT with our optimized version.


\vspace{-0.2cm}
\paragraph{Contributions.} Our contributions are two-fold: \textbf{i)} We mitigate the hallucination problem in IT-VLMs by directly enhancing the grounding of the visual encoder, tuning it to capture fine-grained visual details. We obtain 11.2\% relative improvement in the MMVP benchmark and 6.3\% relative improvement in the MERLIM benchmark \textbf{ii)} EAGLE  is a straightforward and effective approach that reduces hallucinations in IT-VLMs without requiring any additional alignment or tuning. Therefore, \methodname visual modules can be directly integrated into an IT-VLM.   



To ensure reproducibility and to foster future research, upon acceptance, we will make available all the resources related to this paper, including training code, checkpoints, and official benchmark results. 


%



\section{Related Work}

Hallucinations in IT-VLMs, are erroneous or ungrounded elements generated during multimodal inference. These events pose significant challenges to the reliability and trustworthiness of these muti-modal systems. Hallucinations often arise from the model reliance on spurious correlations between visual and textual modalities, inherent language biases within the language model component \cite{zhou2023analyzing}, a tendency to prioritize linguistic cues over visual information \cite{guan2023hallusionbench,li2023evaluating}, and limitations within the visual encoder's capacity to capture fine-grained visual details \cite{Tong_2024_CVPR, villa2024magicmerlimmultimodalevaluation}.

Existing strategies for mitigating IT-VLM hallucinations focus mainly on curating and collecting more instructional data, improving multimodal alignment, and improving visual grounding. 

\vspace{-0.2cm}
\paragraph{Instructional Data Deficiencies.} Prior works \cite{dai2024instructblip, liu2023improvedllava, Qwen-VL} demonstrate that increasing data granularity is crucial for enhancing the instructional capabilities of models while mitigating hallucinations. For example, InstructBLIP \cite{dai2024instructblip} consolidates 13 diverse multimodal datasets spanning tasks like image captioning, reading comprehension from captions, and image-based question answering to strengthen the model's instructional performance.  LLaVA-v1.5 \cite{liu2023improvedllava} refines this approach by selecting four academic datasets from those used by InstructBLIP and transforming them into conversational formats using GPT \cite{openai2023gpt4}, as outlined in \cite{liu2023visual}.

\vspace{-0.2cm}
\paragraph{Adapter module.} The adapter (fusion module) is critical in aligning the two modality encoders \cite{blip}, this adapter learns to prompt the language decoder from the visual embedding and thus an essential step for generating accurate and visually grounded responses. Recent work \cite{blip, dai2024instructblip, liu2023improvedllava} has modified adapter architectures to enhance the model's instructional capabilities. InstructBLIP, an extension of BLIP-2, incorporates the question into its Q-Former module to select the most relevant visual features, enabling more accurate and aligned answers.

\vspace{-0.2cm}
\paragraph{Improving Visual Grounding.} IT-VLMs typically use CLIP-based models as visual encoders \cite{sun2023evaclipimprovedtrainingtechniques, pmlr-v139-radford21a}. However, recent studies \cite{Tong_2024_CVPR, kar2024brave, villa2024magicmerlimmultimodalevaluation} reveal that CLIP-based encoders struggle to capture fine-grained visual details, resulting in visual representations that often fall short for generating grounded and accurate text responses. To address this, works such as \cite{Tong_2024_CVPR, kar2024brave} propose ensembles of visual encoders that leverage the strengths of multiple models to enrich visual representations. While effective, these approaches do not scale well, substantially increasing the computational demand. In contrast, we introduce EAGLE, a training strategy that enhances fine-grained representation in CLIP-based encoders without sacrificing their zero-shot capabilities. Our approach allows for seamless replacement of default visual encoders in IT-VLMs with their EAGLE-tuned counterparts, requiring no architectural modifications, additional encoders, or alignment training. EAGLE demonstrates a significant reduction in hallucinations across 6 state-of-the-art IT-VLMs evaluated on 3 challenging benchmarks.

\begin{figure*}[t]
  \centering
   \includegraphics[width=0.99\linewidth]{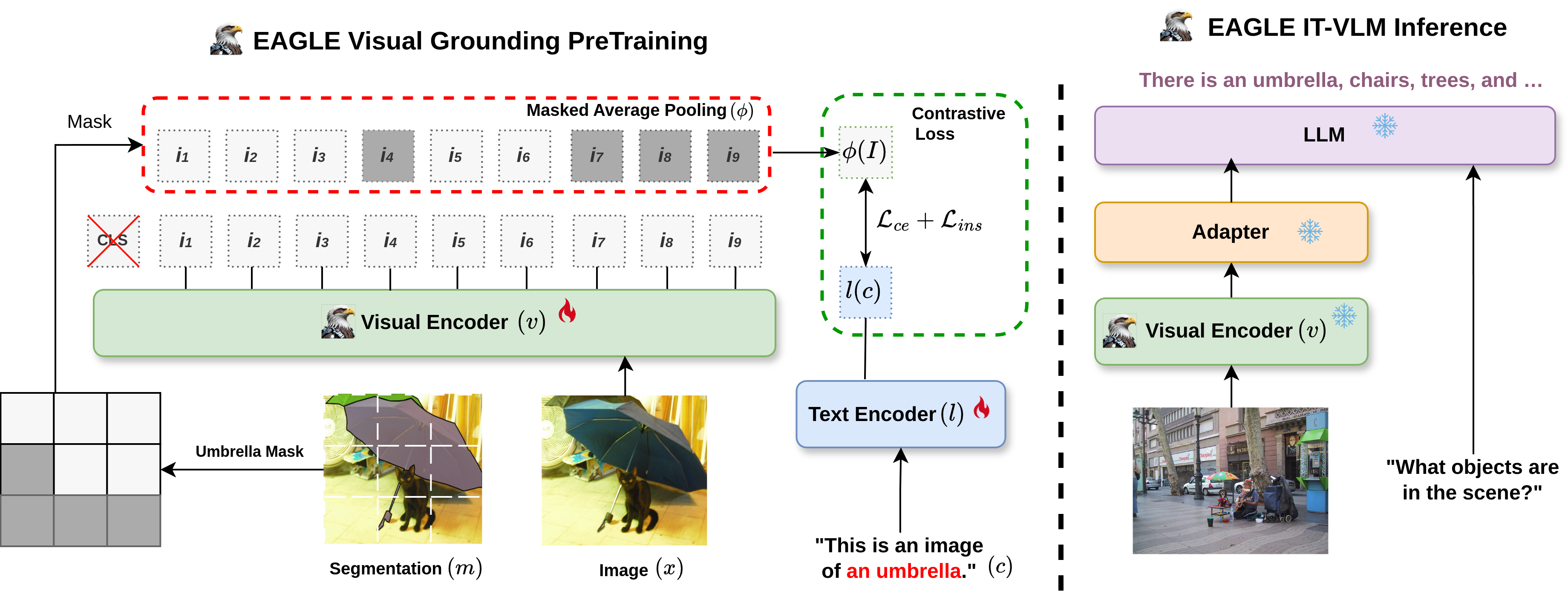}
   \caption{\textbf{Overview of the EAGLE method.} EAGLE reduces the hallucinations in IT-VLMs by improving the grounding of the image encoder. In the post-pretraining phase (Left), EAGLE enhances fine-grained visual representations by employing a masked average pooling (in red dashed lines). This method selects embeddings within the feature sequence corresponding to a specific object and computes an averaged representation. Subsequently, EAGLE enforces local alignment with the language representation of the object (in green dashed lines). The resulting image encoder integrates with any IT-VLM (Right) at inference time, effectively reducing hallucinations without requiring any further tuning.}
   \label{fig:eagle}
\end{figure*}

\section{Methodology}
EAGLE improves the local patch embeddings allowing the ViT to encode refined information about objects in the image using a modified contrastive learning framework. We motivate our strategy by comparing the language alignment of the visual features used in a VLM and an IT-VLM. Both models build upon the same visual encoder, however, VLMs perform alignment with the projection of a single visual token (the CLS token), while IT-VLMs drop the CLS token, and proceed to prompt the language stream with features extracted from the embedded tokens.

\begin{table}[b]
    \caption{\textbf{Zero-Shot performance for VLMs. } When we apply the pipeline of \methodname in a VLM, we observe a small performance degradation in the zero-shot accuracy, as a trade of, we obtain a significant improvement in the alignment of the visual features sequence with the language modality. Therefore, the feature sequence (seq.)  exhibits a significant improvement in a zero-shot evaluation in ImageNet-1K (IN-1K).}\vspace{-0.2cm}
    
    \resizebox{8.5cm}{!}{
        \begin{tabular}{l ll} 
            \toprule
            \multicolumn{1}{c}{\multirow{2}{*}{\bf Model}} & \multicolumn{2}{c}{\bf IN-1K Zero-Shot} \\
            \cmidrule(lr){2-3}
                 & \multicolumn{1}{c}{\bf Cls Acc $\uparrow$} & \multicolumn{1}{c}{\bf Seq. Acc $\uparrow$} \\
            \midrule
            EVA01 ViT-g-14 & \bf 78.50\% & 57.76\% \\
            EAGLE EVA01 ViT-g-14 & 76.65\% \footnotesize\textcolor{red}{(-1.85\%)} & \bf 65.81\% \footnotesize\textcolor{ForestGreen}{(+8.05\%)} \\
            \cmidrule{1-3}
            OpenAI ViT-L-14-336 & \bf 76.55\% & 0.7\% \\
            EAGLE OpenAI ViT-L-14-336 & 71.46\% \footnotesize\textcolor{red}{(-5.09\%)} & \bf 50.98\% \footnotesize\textcolor{ForestGreen}{(+50.28\%)} \\
            \bottomrule
        \end{tabular}
    }
    \label{tab:zst_results_part1}
\end{table}

Table \ref{tab:zst_results_part1} (rows 1 and 3) shows that there is a significant gap in the alignment for the two feature sets. For two representatives VLMs \cite{sun2023evaclipimprovedtrainingtechniques, pmlr-v139-radford21a}, the CLS token is far more aligned with the language embedding and is therefore more effective in a zero-shot recognition task. This result suggests that IT-VLMs are being fine-tuned on sub-optimal sets of visual features. The fusion module is therefore learning how to effectively prompt the language decoder from the visual embeddings, and also correcting for the spurious object representations in the token sequence. For the remainder of this paper, we refer to the input token sequence of the IT-VLM as the “feature sequence". 

As outlined in Figure \ref{fig:eagle}, \methodname ensures that, the ViT module encodes a fine-grained visual representation where individual visual tokens encode a better object representation that remains highly aligned with the language component. In addition, our proposal also preserves the global structure of the feature embedding thus retaining the zero-shot and linear probing capabilities of the original transformer. The \methodname training scheme benefits the visual encoder in two key aspects. First, the visual transformer has almost the same global feature representation and can be readily plugged into any pre-trained IT-VLM. Second, the improved visual representation in the feature sequence will be aware of the object semantics and their spatial alignment in the visual data, thus reducing hallucinations in IT-VLMs.

\subsection{Language Feature Alignment}
\label{sec:eagle_loss} 
To generate a language-aligned representation over the feature sequence of the ViT, we rely on image data with instance level segmentations. Formally, given a trained VLM with visual encoder $v$ and language encoder $l$, we use $v$ to obtain the feature sequence $I$ of image $x$ $I = v(x)$. $I$ has the same sequence length as the tokenized version of $x$ (\textit{i.e} we drop the CLS token). The architecture of $l$ remains unchanged, thus the embedding of the language remains at length 1 (see blue and light gray tokens in figure \ref{fig:eagle}).
 
For a binary segmentation mask $m$, we define a masked average pooling function $\phi(I, m)$ where the feature tokens in $I$ are set to 0 if they do not overlap with the mask $m$, then we perform average pooling. For an image with multiple segmented instances, we randomly sample one of the masks. 

\vspace{-0.2cm}
\paragraph{Loss Function.}
We can not directly use contrastive learning \cite{hadsell2006dimensionality} as our training objective, the standard contrastive loss considers all the elements in the batch as negatives except one. This is incompatible with our setup as the training batch might contain multiple masks with the same semantic class ($c$). 

We decompose our loss into two terms. $\mathcal{L}_{\text{ins}}$ operates at instance level, since there is a single ground-truth class for every mask.  $\mathcal{L}_{\text{ins}}$ is a standard contrastive loss minimizing the distance between the visual features of a single image and the ground-truth language embedding of all the classes. For an image $x$, segmentation mask $m$, and mask's class $c$, we obtain a pair of tokens $\{\phi(v(x), m), l(c)\}$ and define the instance level loss as:

\begin{equation}
    \mathcal{L}_{\text{ins}} = \mathcal{L_{\text{con}}}(\phi(I, m), l(c_{j})),
\end{equation}

where $\mathcal L_{\text{con}}$ is the contrastive loss. Note that $c_{j}$ is composed of a prefix and the class ($j$) of a mask in natural language, $c_j$ follows a prompt template where we prepend “This is an image of $<$mask\_class$>$". 

Our second loss function $\mathcal{L}_{\text{ce}}$ allows for multiple masks to be associated to a single class. To this end, we leverage a cross-entropy loss over a distance measure ($d_j$) between the modalities, normalized with a sigmoid function.

\begin{table*}[t]
    
    \centering
     \caption{
        \textbf{Results on MMVP-VLM Benchmark.} We evaluate the performance (Accuracy) of EAGLE-tuned CLIP models across the nine tasks defined in the MMVP-VLM Benchmark. EAGLE demonstrates consistent performance improvements on most tasks, particularly when using the feature sequence representations (SEQ). Even though the CLS token (CLS) is not directly trained in EAGLE, the tuning process still yields a performance boost for the CLS token in the EVA01 ViT model. The tasks in MMVP-VLM are: \faCompass\ Orientation and Direction. \faSync\ State and Condition. \faSortNumericUp\ Quantity and Count. \faMapPin\ Positional and Relational Context. \faPalette\ Color and Appearance. \faCogs\ Structural and Physical Characteristics. \faFont\ Text. \faCamera\ Viewpoint and Perspective.
    }\vspace{-0.2cm}
    
    \resizebox{17.5cm}{!}{
    \begin{tabular}{ll rrrrrrrrrl} 
        \toprule
        
        \multicolumn{1}{c}{\bf VLM} & \multicolumn{1}{c}{\bf Feature} &  \multicolumn{1}{c}{\bf \faCompass} & \multicolumn{1}{c}{\bf \faSearch} & 
        \multicolumn{1}{c}{\bf \faSync} &  \multicolumn{1}{c}{\bf \faSortNumericUp} & \multicolumn{1}{c}{\bf \faMapPin} & \multicolumn{1}{c}{\bf \faPalette} &  \multicolumn{1}{c}{\bf \faCogs} & \multicolumn{1}{c}{\bf \faFont} & 
        \multicolumn{1}{c}{\bf \faCamera} & 
        \multicolumn{1}{c}{\bf Average} 
        \\\cmidrule(lr){1-12}
        EVA-01-CLIP-g-14 & CLS & 6.67\% & \bf 26.67\% & \bf 53.33\% & \bf 6.67\% & \bf 13.33\% & 53.33\%  & \bf 26.67\% & 6.67\% & \bf 33.33\% & 25.18\%
        \\
        EAGLE EVA-01-CLIP-g-14 & CLS & \bf 13.33\% & 20.0\% & \bf 53.33\% & \bf 6.67\% & \bf 13.33\% & \bf 66.67\% & \bf 26.67\% & \bf 13.33\%  & 26.67\% & \bf 26.67\% \footnotesize \textcolor{ForestGreen}{(-1.49\%)}
        \\
        EVA-01-CLIP-g-14 & SEQ & 0.0\% & \bf 26.67\% & \bf 46.67\% & \bf 13.33\% & 0.0\% & 33.33\% & 13.33\%  & 0.0\%  & \bf 20.0\% & 17.04\%
        \\
        EAGLE EVA-01-CLIP-g-14 & SEQ & \bf 6.67\% & 20.0\% & \bf 46.67\% &  6.67\% & \bf 6.67\% & \bf 46.67\%  & \bf 20.0\% & \bf 13.33\% & 13.33\% & \bf 20.00\% \footnotesize \textcolor{ForestGreen}{(+2.96\%)}
        \\\cmidrule(lr){1-12}
        OpenAI CLIP-L-14-336 & CLS & \bf 0.0\% & \bf 20.0\% & \bf 40.0\% & \bf 20.0\% & \bf 6.67\% & \bf 20.0\% & \bf 33.33\% & 6.67\% & \bf 26.67\% & \bf 19.26\%
        \\
         EAGLE OpenAI CLIP-L-14-336 & CLS & 0.0\% & 20.0\% & 33.33\% & 6.67\% & 0.0\% & \bf 20.0\% & 26.67\% & \bf 13.33\% & 20.0\% & 15.56\% \bf \footnotesize \textcolor{red}{(-3.70\%)}
        \\
        OpenAI CLIP-L-14-336 & SEQ & \bf 13.33\% & 6.67\% & 0.0\% & \bf 13.33\% & 0.0\% & 26.67\% & 13.33\% & 6.67\% & 0.0\% & 8.89\%
        \\
         EAGLE OpenAI CLIP-L-14-336 & SEQ & \bf 20.0\% & \bf 33.33\% & \bf 33.33\% & 0.0\% & \bf 6.67\% & \bf 60.0\% & \bf 20.0\% & \bf 6.67\% & \bf 20.0\% & \bf 22.22\% \footnotesize \textcolor{ForestGreen}{(+13.33\%)}
        \\
        \bottomrule
    \end{tabular}
}
   
    \label{tab:mmvp_results}
\end{table*}

\begin{align}
    d_{j} &= 1 - \sigma (\phi(I, m) -  l(c_{j})) \\
    \mathcal{L}_{ce} &= -\frac{1}{K} \sum_{j=1}^K c_j \log(d_j) + (1 - c_j) \log(1 - d_j)
\end{align}
where $\sigma$ is the sigmoid function, and $K$ is the total number of classes in the training set. Our final loss function integrates both losses $\mathcal{L}_{\text{ce}}$ and $\mathcal{L}_{\text{ins}}$ without extra hyperparameters.

\begin{equation}
    \mathcal{L} = \mathcal{L}_{\text{ce}} +  \mathcal{L}_\text{ins}
\end{equation}
Empirically, we find that it is beneficial to have an approximate uniform distribution of training classes. Therefore, we resample such that masks with uncommon classes are more likely to be sampled than masks belonging to the most common classes.

\vspace{-0.2cm}
\paragraph{Training Dataset.} 
EAGLE requires image data with instance segmentation labels. We leverage the labeled data in the OpenImages V7 dataset \cite{openimagesv7}, which includes 944,037 images annotated with object segmentations distributed over 350 object classes. Although the data available in the segmented set of OpenImages is three orders of magnitude smaller than the datasets used to pre-train VLM models \cite{sun2023evaclipimprovedtrainingtechniques, pmlr-v139-radford21a}, we show that the  training strategy in EAGLE suffices to learn fine-grained visual representations at the patch level that are directly transferable to IT-VLMs. It is also important to note that OpenImages V7 lacks any instructional data, or training data for VQA tasks, therefore \methodname does not represent any in-domain training for IT-VLMs.

\begin{table}[b]
    \caption{\textbf{EAGLE on VLMs.} We apply the eagle pipeline to two standard VLMS, and fine-tune the model with data from the OpenImagesV7 dataset, we observe a significant improvement in the false positive ratio (FP) when evaluating in the MS-COCO dataset.} \vspace{-0.2cm}
    
    \resizebox{8.5cm}{!}{
        \begin{tabular}{l lll} 
            \toprule
            \multicolumn{1}{c}{\multirow{2}{*}{\bf Model}} & \multicolumn{3}{c}{\bf MS-COCO False Positives} \\
            \cmidrule(lr){2-4}
                 & \multicolumn{1}{c}{\bf Seq. FP@1 $\downarrow$} & \multicolumn{1}{c}{\bf Seq. FP@3 $\downarrow$} & \multicolumn{1}{c}{\bf Seq. FP@5 $\downarrow$} \\
            \midrule
            EVA01 ViT-g-14 & 24.64\% & 56.68\% & 68.17\% \\
            EAGLE EVA01 ViT-g-14 & \bf 5.11\% \footnotesize\textcolor{ForestGreen}{(-19.53\%)} & \bf 39.57\% \footnotesize\textcolor{ForestGreen}{(-17.11\%)} & \bf 57.77\% \footnotesize\textcolor{ForestGreen}{(-10.4\%)} \\
            \cmidrule{1-4}
            OpenAI ViT-L-14-336 & 43.74\% & 67.63\% & 77.09\% \\
            EAGLE OpenAI ViT-L-14-336 & \bf 29.54\% \footnotesize\textcolor{ForestGreen}{(-14.2\%)} & \bf 51.35\% \footnotesize\textcolor{ForestGreen}{(-16.28\%)} & \bf 63.64\% \footnotesize\textcolor{ForestGreen}{(-13.45\%)} \\
            \bottomrule
        \end{tabular}
    }
    \label{tab:zst_results_part2}
\end{table}

\vspace{-0.2cm}
\paragraph{Parameter Efficient Fine-Tuning.} Empirically, we observe that our loss function enhances the visual grounding of $v$, but introduces substantial drift from the original feature distribution of the pre-trained VLM. This drift causes significant performance drops in zero-shot accuracy (see table \ref{tab:zst_results_part1}) after fine-tuning. To address this, we incorporate Gradient Low-Rank Projection (GaLore) \cite{zhao2024galore} into our training pipeline. GaLore is a memory-efficient fine-tuning technique that facilitates full-parameter learning by computing a low-rank approximation of the gradient $G$ for each weight matrix $W$. This method is more memory-efficient than directly optimizing rank-decomposed weight matrices, as in LoRA \cite{hu2021lora}. Although GaLore supports optimization of all parameters, we follow prior work \cite{hu2021lora, zhao2024galore, yang2024corda} by focusing on the MLPs and linear layers within the attention-based architecture. These parameters and memory-efficient fine-tuning techniques have demonstrated effectiveness in mitigating distribution shift and catastrophic forgetting in continual learning scenarios \cite{gao2023unified, villa2023pivot}, enabling better retention of the model’s original performance on pre-training tasks.



\begin{table*}[!h]
    \centering
    
    \caption{
        \textbf{Results on POPE and MMVP Benchmarks.} We assess the effectiveness of the EAGLE-tuned visual encoder on multiple state-of-the-art IT-VLMs on the POPE and MMVP benchmarks. Without any tuning or alignment to the LLM or its fusion model, EAGLE consistently enhances performance across all of the IT-VLMS for all metrics in both benchmarks. LLaVA-v1.5*, the only version incorporating alignment training, demonstrates the best performance gains on the MMVP benchmark.  
    } \vspace{-0.2cm}
    
    \resizebox{15cm}{!}{
    \begin{tabular}{l l l rrrr |r} 
        \toprule
        \multicolumn{1}{c}{\multirow{2}{*}{\bf Model}} &  \multicolumn{1}{c}{\multirow{2}{*}{\bf Visual Encoder}} & \multicolumn{1}{c}{\multirow{2}{*}{\bf LLM}} & \multicolumn{4}{c}{\bf POPE} & \multicolumn{1}{c}{\bf MMVP}
        \\\cmidrule(lr){4-8}
             & & & \multicolumn{1}{c}{$ \bf Acc \uparrow$} & \multicolumn{1}{c}{$ \bf F1 \uparrow$} & \multicolumn{1}{c}{$\bf Prec \uparrow $} & \multicolumn{1}{c}{$\bf Rec \uparrow$} & \multicolumn{1}{c}{$\bf Acc \uparrow$} 
             \\
        \midrule
        MiniGPT-4 & EVA01 ViT-g-14 & Vicuna-7B v0 & 55.13\% & 64.44\% & 53.36\% & 81.33\% & 11.33\%
        \\
        MiniGPT-4 & \textbf{EAGLE} EVA01 ViT-g-14 & Vicuna-7B v0 & \bf 55.67\% & \bf 64.96\% & \bf 53.70\% & \bf 82.2\% & \bf 14.67\%
        \\
        \cmidrule{1-8}
        BLIP-2 & EVA01 ViT-g-14 & FlanT5xl & 69.33\% & 70.90\% & 67.34\% & 75.06\% & 14.0\% 
        \\
        BLIP-2 & \textbf{EAGLE} EVA01 ViT-g-14 & FlanT5xl & \bf 71.40\% & \bf 72.81\% & \bf 69.40\% & \bf 76.60\% & \bf 15.33\%
        \\
        \cmidrule{1-8}
        InstructBLIP & EVA01 ViT-g-14 & Vicuna-7B v1.1 & 75.0\% & 77.99\% & 69.73\% & \bf 88.47\% & 18.0\% 
        \\
        InstructBLIP & \textbf{EAGLE} EVA01 ViT-g-14 & Vicuna-7B v1.1 & \bf 77.20\% & \bf 79.39\% & \bf 72.44\% & 87.80\% & \bf 20.67\%
        \\
        InstructBLIP & EVA01 ViT-g-14 & Vicuna-13B v1.1 & 64.43\% & 73.73\% & 59.46\% & \bf 97.0\% & \bf 24.67\%
        \\
        InstructBLIP & \textbf{EAGLE} EVA01 ViT-g-14 & Vicuna-13B v1.1 & \bf 67.57\% & \bf 74.92\% & \bf 61.08\% & 96.8\% & \bf 24.67\%
        \\
        InstructBLIP & EVA01 ViT-g-14 & FlanT5xl & 58.0\% & 70.0\% & 54.44\% & \bf 98.0\% & 18.0\%
        \\
        InstructBLIP & \textbf{EAGLE} EVA01 ViT-g-14 & FlanT5xl & \bf 59.83\% & \bf 70.92\% & \bf 55.58\% & 97.93 & \bf 19.33\%
        \\
        \cmidrule{1-8}
        LLaVA-v1.5 & OpenAI ViT-L-14-336 & Vicuna-7B v1.5 & \bf 69.33\% & \bf 76.10\% & \bf 62.34\% & 97.67\% & 25.33\%
        \\
         LLaVA-v1.5 & \textbf{EAGLE} OpenAI ViT-L-14-336 & Vicuna-7B v1.5 & 67.53\% & 74.93\% & 61.02\% & 97.07\% & 26.66\%
        \\
        LLaVA-v1.5* & \textbf{EAGLE} OpenAI ViT-L-14-336 & Vicuna-7B v1.5 & 68.26\% & 75.60\% & 61.41\% & \bf 98.33\% & \bf 30.67\%
        \\
        \bottomrule
    \end{tabular}
    }
    \label{tab:itlvlm_results_part1}
\end{table*}

\section{Experimental Evaluation}
\label{sec:exp_sec}



Before we check the capabilities of ViTs enhanced with EAGLE in the instructional domain of IT-VLMs, we test the effectiveness of our method in the VLM domain. We fine-tune $v$ and $l$ from the pre-trained EVA-01-CLIP-g-14 and OpenAI CLIP-L-14-336 VLMs \cite{pmlr-v139-radford21a, sun2023evaclipimprovedtrainingtechniques} and verify if their feature sequence is any more effective at recognizing multiple objects in an image. We validate our VLMs in the MS-COCO dataset \cite{lin2014microsoft}, and check for improvements regarding the number of False Positives predictions. Our testing methodology follows the zero-shot protocol of \cite{sun2023evaclipimprovedtrainingtechniques}, and we rank the top K predictions. Within this rank, we calculate the ratio of false positives.

Table \ref{tab:zst_results_part2} summarizes the results of our preliminary evaluation on VLMs. EAGLE ViTs significantly reduce the number of false positive predictions in VLMs, with up to 19.53\% less false positive errors.

\vspace{-0.3cm}
\paragraph{Implementation Details.} We tune the VLMs EVA01-CLIP-g-14 \cite{sun2023evaclipimprovedtrainingtechniques} and OpenAI CLIP-L-14-336 \cite{pmlr-v139-radford21a}. The ViT in these models corresponds to the visual module in MiniGPT-4 \cite{zhu2023minigpt}, BLIP-2 \cite{blip}, InstructBLIP \cite{dai2024instructblip}, and LLaVA-v1.5 \cite{liu2023improvedllava}. We apply GaLore \cite{zhao2024galore}, setting the rank to 128 and the learning rate to 4e-6. The training is performed with two A100 GPUs (80GB) with a batch size of 512. We train until convergence of the $\mathcal{L}_{m}$ loss. EVA01-CLIP-g-14 and OpenAI CLIP-L-14-336 converge at epoch 8, and training requires about 20 hours.

\subsection{Visual Representation Quality} 

We now proceed with an in-depth empirical evaluation testing the EAGLE-tuned VLMs in the MMVP-VLM benchmark \cite{Tong_2024_CVPR}. MMVP-VLM designs nine challenging scenarios where CLIP-based models typically fail. The failure case is designed by pairing visually distinct images that have a highly similar CLIP feature embedding. As a consequence, VLMs fail to align the textual description of those image pairs. MMVP-VLM characterizes nine possible scenarios for the visual changes in the image content:  Orientation and Direction ({\small \faCompass}),  Presence of Specific Features ({\small \faSearch}),  State and Condition ({\small \faSync}),  Quantity and Count ({\small \faSortNumericUp}),   Positional and Relational Context ({\small \faMapPin}), Color and Appearance ({\small \faPalette}),  Structural and Physical Characteristics ({\small \faCogs}), and   Text ({\small \faFont}).

Table \ref{tab:mmvp_results} presents the results of EAGLE-tuned versions of OpenAI CLIP-L-14-336 and EVA-01-CLIP-g-14 models in MMVP. While MMVP remains a highly challenging benchmark, EAGLE demonstrates clear improvements across most of the evaluated scenarios. In particular, the EVA-01-CLIP-g-14 model reports improvement for features extracted from the CLS token and the feature sequence. We emphasize that EAGLE’s tuning strategy does not update the CLS token. Despite this, the CLS token in EVA-01-CLIP-g-14 ViT exhibits an average performance boost of 1.49\%. For OpenAI CLIP-L-14-336, EAGLE tuning significantly enhances the performance of the feature sequence, even surpassing the CLS token's average performance. These results further verify EAGLE's effectiveness in enhancing the model's ability to capture fine-grained visual details.

\subsection{Hallucinations in Instructional Models}
We now proceed with a comprehensive empirical evaluation of state-of-the-art IT-VLMs enhanced with EAGLE on multi-modal instructional benchmarks. We select three representative hallucination benchmarks: POPE \cite{li2023evaluating}, MMVP \cite{Tong_2024_CVPR}, and MERLIM \cite{villa2024magicmerlimmultimodalevaluation}. We run these benchmarks on six different IT-VLMs which are augmented with EAGLE-tuned visual encoders.




\vspace{-0.3cm}
\paragraph{Hallucination Benchmarks.}

POPE \cite{li2023evaluating} assesses hallucination events using yes/no questions about objects in an image. For this benchmark, we focus on POPE's most challenging subset: Adversarial SEEM \cite{zou2024segment} from A-OKVQA \cite{aokvqa}, which uses the SEEM segmentation model to detect object segmentations in A-OKVQA images. In POPE, questions with “yes” answers are generated based on ground truth objects, while questions with “no” answers are formulated from the top-k most frequent objects in the dataset, which are not present in the image. POPE is a direct tool to evaluate if the visual embeddings generated by \methodname perform better with queries addressing individual objects instead of the image's global appearance.

MMVP \cite{Tong_2024_CVPR} measures hallucinations using 150 image pairs and 300 corresponding (a) or (b) questions. The image pairs are designed such that they have highly similar CLIP embeddings. MMVP scores favorably only if both questions for each image pair are answered correctly. Following MMVP's methodology, we used a GPT-based scoring approach, replacing the now deprecated “GPT-4-0314” with its closest current successor, “GPT-4o” \cite{openai2023gpt4}. MMVP offers a direct way to evaluate whether or not EAGLE improves the model's recognition of fine-grained visual details.

\begin{table*}[t]
    \centering
    \caption{
        \textbf{Precision of IT-LVLMs on MERLIM Benchmark.} We evaluate the precision of IT-VLMs across semantically identical question pairs on both original and edited image sets within the MERLIM benchmark. Prompt 1 and Prompt 2 are “List the objects that appear in the image.” and “Enumerate the items visible in the picture.”, respectively. Our analysis emphasizes the subset of MERLIM where an entire object category has been removed from the edited images. Notably, EAGLE demonstrates a consistent, significant improvement without multimodal alignment training across all methods and for both image set and question variations. LLaVA-v1.5* represents the version trained to improve the LLM alignment with the EAGLE-tuned visual encoder, which shows even stronger results.
    }\vspace{-0.2cm}
    
    \resizebox{17.5cm}{!}{
    \begin{tabular}{l l l rrrrl} 
        \toprule
        \multicolumn{1}{c}{\multirow{3}{*}{\bf Model}} &  \multicolumn{1}{c}{\multirow{3}{*}{\bf Visual Encoder}} & \multicolumn{1}{c}{\multirow{3}{*}{\bf LLM}} & \multicolumn{5}{c}{\bf MERLIM}
        \\\cmidrule(lr){4-8}
         & & & \multicolumn{2}{c}{\multirow{1}{*}{\bf Prompt 1}} & \multicolumn{2}{c}{\multirow{1}{*}{\bf Prompt 2}} & \multicolumn{1}{c}{\multirow{2}{*}{$\bf Avg \uparrow$}}
        \\\cmidrule(lr){4-7}
             & & & \multicolumn{1}{c}{$\bf Prec_{Orig} \uparrow$} & \multicolumn{1}{c}{$\bf Prec_{Inp} \uparrow$} & \multicolumn{1}{c}{$\bf Prec_{Orig} \uparrow$} & \multicolumn{1}{c}{$\bf Prec_{Inp} \uparrow$} & 
             \\
        \midrule
        MiniGPT-4 & EVA01 ViT-g-14 & Vicuna-7B v0 & 36.68\% & 31.28\% & 37.68\% & 31.93\% & 34.39\%
        \\
        MiniGPT-4 & EAGLE EVA01 ViT-g-14 & Vicuna-7B v0 & \bf 37.55\% & \bf 32.73\% & \bf 38.91\% & \bf 33.38\% & \bf 35.64\% \footnotesize\textcolor{ForestGreen}{(+1.25\%)}
        \\
        \cmidrule{1-8}
        BLIP-2 & EVA01 ViT-g-14 & FlanT5xl & 57.23\% & 46.84\% & 58.86\% & 47.47\% & 52.60\%
        \\
        BLIP-2 & EAGLE EVA01 ViT-g-14 & FlanT5xl & \bf 59.95\% & \bf 49.58\% & \bf 63.06\% & \bf 50.73\% & \bf 55.83\% \footnotesize\textcolor{ForestGreen}{(+3.23\%)}
        \\
        \cmidrule{1-8}
        InstructBLIP & EVA01 ViT-g-14 & Vicuna-7B v1.1 & 57.92\% & 48.47\% & 44.45\% & 39.28\% & 47.53\% 
        \\
        InstructBLIP & EAGLE EVA01 ViT-g-14 & Vicuna-7B v1.1 & \bf 60.90\% & \bf 50.90\% & \bf 47.42\% & \bf 41.86\% & \bf 50.27\% \footnotesize\textcolor{ForestGreen}{(+2.74\%)}
        \\
        InstructBLIP & EVA01 ViT-g-14 & Vicuna-13B v1.1 & 36.50\% & 32.02\% & 32.44\% & 28.43\% & 32.35\%
        \\
        InstructBLIP & EAGLE EVA01 ViT-g-14 & Vicuna-13B v1.1 & \bf 40.67\% & \bf 35.35\% & \bf 35.28\% & \bf 31.39\% & \bf 35.67\% \footnotesize\textcolor{ForestGreen}{(+3.32\%)}
        \\
        InstructBLIP & EVA01 ViT-g-14 & FlanT5xl & 41.30\% & 36.71\% & 44.13\% & 38.52\% & 40.17\%
        \\
        InstructBLIP & EAGLE EVA01 ViT-g-14 & FlanT5xl & \bf 43.58\% & \bf 38.78\% & \bf 47.76\% & \bf 42.33\% & \bf 43.11\% \footnotesize\textcolor{ForestGreen}{(+2.94\%)}
        \\
        \cmidrule{1-8}
        LLaVA-1.5 & OpenAI ViT-L-14-336 & Vicuna-7B v1.5 &  49.52\% & 42.57\% & 30.41\% & 28.24\% & 37.69\%
        \\
         LLaVA-1.5 & EAGLE OpenAI ViT-L-14-336 & Vicuna-7B v1.5 & \bf 50.34\% & \bf 43.03\% & 31.10\% & 28.32\% & 38.20\% \bf \footnotesize\textcolor{ForestGreen}{(+0.51\%)}
        \\
        LLaVA-1.5* & EAGLE OpenAI ViT-L-14-336 & Vicuna-7B v1.5 & 48.85\% & 42.33\% & \bf 38.24\% & \bf 33.08\% & \bf 40.62\% \footnotesize\textcolor{ForestGreen}{(+2.93\%)}
        \\
        \bottomrule
    \end{tabular}
    }
    \label{tab:itlvlm_results_part2}
\end{table*}

In MERLIM \cite{villa2024magicmerlimmultimodalevaluation}, we evaluate EAGLE using a subset of original and edited images. We target those images where an entire object category was removed from the edited image, \textit{i.e.} there exists only 1 instance of the object in the image and was removed. This subset results in 5608 edited images and 3037 corresponding originals. MERLIM incorporates open-ended questions with equivalent meanings, to inquire about all the objects present in the image. This design choice provides a more realistic scenario to assess both EAGLE's ability to capture fine-grained visual information of individual objects and the impact of the prompt phrasing.

\vspace{-0.3cm}
\paragraph{Results on Hallucionation Benchmarks.} EAGLE enhances the performance of all evaluated IT-VLMs across the three selected benchmarks without requiring any alignment or tuning of the IT-VLM. We summarize these results in Tables \ref{tab:itlvlm_results_part1} (POPE and MMVP) and \ref{tab:itlvlm_results_part2} (MERLIM). Notably, EAGLE shows a larger improvement in the more challenging benchmarks MMVP (2.3\% absolute improvement and 11.2\% relative improvement) and MERLIM (2.73\% absolute improvement and 6.3\% relative improvement). MMVP requires the model to detect subtle visual variations within image pairs and then correctly align both images with the text prompts. MERLIM checks all nouns (and potential synonyms) in the responses and tests the model on altered images with subtle but meaningful modifications of the original image. These performance improvements highlight EAGLE's capability to encode fine-grained visual information at the feature sequence, while preserving the initial representation space and reducing hallucinations without additional training.

EAGLE achieves larger improvements for BLIP-2 and InstructBLIP, likely due to two factors: (1) EVA01 ViT-g-14 captures better fine-grained visual information at the patch level more effectively than OpenAI’s ViT-L-14-336, as demonstrated in Table \ref{tab:zst_results_part2}. As a consequence, the model EVA01 requires less supervision for $\mathcal{L}_{m}$ to converge, and thus the global representation drifts less in comparison to OpenAI CLIP.  (2) BLIP-2 and InstructBLIP train only the multimodal adapter during the instruction tuning, while LLaVA-v1.5 (using OpenAI ViT-L-14-336) trains both the adapter and the LLM, making it more challenging for LLaVA-v1.5 to fully leverage EAGLE’s optimized visual encoder. Nevertheless, we show that \methodname tuned models can be directly plugged into LLaVA-v1.5 and can improve its performance in MERLIM and MMVP, while retaining 98.5\% of its original performance in POPE.


\vspace{-0.3cm}
\paragraph{Hidden Hallucinations.} The MERLIM benchmark introduces the concept of ``hidden hallucinations'' to describe seemingly correct text responses that remain unchanged after removing their visual grounding. The authors suggest that these answers are probably the result of the LLM ignoring visual data and simply generating consistent text \cite{villa2024magicmerlimmultimodalevaluation}. We assess the impact of \methodname on reducing hidden hallucinations. As shown in Table \ref{tab:itlvlm_results_part2}, EAGLE achieves significant and consistent performance gains on both original (Orig.) and edited images (Inp.). These results evidence the effectiveness of \methodname at capturing fine-grained visual details, but also at providing more effective visual features that allow the IT-VLMs to rely less on previously generated language tokens to produce an answer. 

\begin{table*}[ht]
    \caption{\textbf{Training Strategy Analysis.} We investigate the effects of EAGLE’s training strategy and supervision. Notably, GaLore enables EAGLE to preserve zero-shot performance while enhancing fine-grained visual representation within the sequence embeddings, as evidenced by a substantial reduction in the false positive rate on MS-COCO.} \vspace{-0.2cm}

    \centering
    \resizebox{15cm}{!}{
        \begin{tabular}{lllrrrr} 
            \toprule
            \multicolumn{1}{c}{\multirow{2}{*}{\bf Visual Model}} & \multicolumn{1}{c}{\multirow{2}{*}{\bf Training}} & \multicolumn{1}{c}{\multirow{2}{*}{\bf Supervision}} & \multicolumn{2}{c}{\bf IN-1K ZS} & \multicolumn{2}{c}{\bf MS-COCO False Positives} \\
            \cmidrule(lr){4-7}
                 & & & \multicolumn{1}{c}{\bf Cls Acc $\uparrow$} & \multicolumn{1}{c}{\bf Seq. Acc $\uparrow$} & \multicolumn{1}{c}{\bf Seq. FP@1 $\downarrow$} & \multicolumn{1}{c}{\bf Seq. FP@3 $\downarrow$}\\
            \midrule
            EVA01 ViT-g-14 (Baseline) & None & None & \textbf{78.50\%} & 57.76\% & 24.64\% & 56.68\% \\
            EVA01 ViT-g-14 & Full-Finetune & CLS & 70.90\% & 44.55\% & 17.87\% & 51.57\% \\
            EVA01 ViT-g-14 & Full-Finetune & Seq. & 71.88\% & 61.21\% & \bf 3.75\% & \bf 38.02\% \\
            EVA01 ViT-g-14 & GaLore & CLS \& Seq. & 74.24\% & 61.66\% & 4.62\% & 38.14\% \\
            EAGLE EVA01 ViT-g-14 & GaLore & Seq. & 76.65\% & \textbf{65.81\%} & 5.11\% & 39.57\% \\
            \bottomrule
        \end{tabular}
    }
    \label{tab:ablations}
\end{table*}

\vspace{-0.3cm}
\paragraph{Prompt Bias.} 
We continue our analysis on hallucination by briefly analyzing the instruction bias in IT-VLMs. The MERLIM benchmark has multiple instructions for a single task, thus allowing us to analyze the effect of syntactically different instructions with the same semantic meaning. We observe that our EAGLE pre-trained models do not alter the instruction bias. Table \ref{tab:itlvlm_results_part2}  shows that \methodname always increases the precision in both prompts, but never switches their relative performance. That is, if prompt 1 was better than prompt 2 with the original ViT, this pattern persists after switching to our \methodname tuned encoder.


\vspace{-0.3cm}
\paragraph{LLaVA Training.} The LLaVA-v1.5 model is a special case among the IT-VLMs. During the instructional tuning stage, both the adapter module and the language model (LLM) are tuned. This setup introduces additional challenges for the transfer of our \methodname models, as the tuned adapter and LLM might not directly interface with the EAGLE-enhanced feature representation. To better analyze this feature shift, we apply our \methodname visual encoder as is, and we also reproduce the instructional tuning stage of LLaVA-v1.5, using our \methodname visual encoder instead of the original OpenAI CLIP-L-14-336 visual stream. Since we train only the second stage, we use the adapter checkpoint obtained from the first stage (alignment stage), and the original Vicuna v1.5 \cite{vicuna2023}. We optimize the ensemble with the the original LLaVA-v1.5 codebase over two A100 GPUs (80GB).

\vspace{-0.4cm}
\paragraph{Impact of Instructional Tuning.} As shown in Tables \ref{tab:itlvlm_results_part1} and \ref{tab:itlvlm_results_part2}, our retrained model (LLaVA-v1.5*) surpasses all IT-VLMs on the MMVP benchmark, achieving a performance gain of 5.34\%, even exceeding the improvement reported on \cite{Tong_2024_CVPR} of 3.3\%. This suggests that   \methodname backbones are highly compatible with instructional pre-training. Moreover, when trained on the downstream task, \methodname overcomes the performance of a far larger visual ensemble that incorporates CLIP and DINOv2 \cite{oquab2023dinov2} representations. This result emphasizes EAGLE’s ability to effectively encode fine-grained visual representations within patch embeddings. Likewise, LLaVA-v1.5* achieves an average improvement of 2.93\% on the MERLIM benchmark, demonstrating that \methodname can improve results across different datasets and benchmarks with the very same instructional training.



\subsection{Training Strategy Analysis}

We conclude the empirical evaluation of EAGLE and examine the contribution of the key design choices in EAGLE’s. 


As outlined in Table \ref{tab:ablations}, the GaLore component is the most important one for maintaining zero-shot accuracy on ImageNet-1K while also enabling the sequence embeddings to capture detailed visual features, as evidenced by a substantial reduction in the false positive rate on MS-COCO. A second important design choice is to refrain from training the CLS token in favor of focusing exclusively on fine-tuning the feature sequence. Although training the CLS token further reduces the false positive rate of sequence embeddings, it compromises the zero-shot performance and reduces the generalization capability of the new feature representation. 



\section{Conclusions and Limitations}
This paper introduces EAGLE, a straightforward and scalable approach for reducing hallucinations in IT-VLMs. Unlike traditional methods that rely on additional instructional data, fine-tuning the LLM component, enhancing the adapter module, or ensembling multiple visual encoders to improve visual representation, EAGLE employs a tuning strategy to enhance fine-grained visual grounding directly within the visual encoder. We demonstrate that EAGLE-tuned visual encoders can integrate seamlessly into an IT-VLM without requiring additional alignment or instructional training. EAGLE improves hallucination metrics across three standard benchmarks and six diverse architectures, which is also consistent with the significant improvements in the challenging scenarios for VLM proposed in the MMVP-VLM benchmark.

\paragraph{Limitations.} EAGLE utilizes object segmentations to encourage fine-grained object representation within the sequence embeddings. However, the relative scarcity of image data with instance segmentations poses a challenge. To address this, we employ a parameter-efficient fine-tuning strategy, such as GaLore, which helps prevent model overfitting and preserves generalization capabilities. We encourage future work to explore surrogate sources of fine-grained supervision to further improve model grounding and performance.
{
    \small
    \bibliographystyle{ieeenat_fullname}
    \bibliography{main}

\begin{thebibliography}{43}
\providecommand{\natexlab}[1]{#1}
\providecommand{\url}[1]{\texttt{#1}}
\expandafter\ifx\csname urlstyle\endcsname\relax
  \providecommand{\doi}[1]{doi: #1}\else
  \providecommand{\doi}{doi: \begingroup \urlstyle{rm}\Url}\fi

\bibitem[Bai et~al.(2023)Bai, Bai, Yang, Wang, Tan, Wang, Lin, Zhou, and Zhou]{Qwen-VL}
Jinze Bai, Shuai Bai, Shusheng Yang, Shijie Wang, Sinan Tan, Peng Wang, Junyang Lin, Chang Zhou, and Jingren Zhou.
\newblock Qwen-vl: A versatile vision-language model for understanding, localization, text reading, and beyond.
\newblock \emph{arXiv preprint arXiv:2308.12966}, 2023.

\bibitem[Benenson and Ferrari(2022)]{openimagesv7}
Rodrigo Benenson and Vittorio Ferrari.
\newblock From colouring-in to pointillism: revisiting semantic segmentation supervision, 2022.

\bibitem[Brown et~al.(2020)Brown, Mann, Ryder, Subbiah, Kaplan, Dhariwal, Neelakantan, Shyam, Sastry, Askell, et~al.]{brown2020language}
Tom Brown, Benjamin Mann, Nick Ryder, Melanie Subbiah, Jared~D Kaplan, Prafulla Dhariwal, Arvind Neelakantan, Pranav Shyam, Girish Sastry, Amanda Askell, et~al.
\newblock Language models are few-shot learners.
\newblock \emph{Adv. Neural Inform. Process. Syst.}, 33:\penalty0 1877--1901, 2020.

\bibitem[Chen et~al.(2020)Chen, Kornblith, Norouzi, and Hinton]{pmlr-v119-chen20j}
Ting Chen, Simon Kornblith, Mohammad Norouzi, and Geoffrey Hinton.
\newblock A simple framework for contrastive learning of visual representations.
\newblock In \emph{Proceedings of the 37th International Conference on Machine Learning}, pages 1597--1607. PMLR, 2020.

\bibitem[Chen and Wang(2022)]{chen2022pali}
Xi Chen and Xiao Wang.
\newblock Pali: Scaling language-image learning in 100+ languages.
\newblock In \emph{Adv. Neural Inform. Process. Syst.}, 2022.

\bibitem[Cherti et~al.(2023)Cherti, Beaumont, Wightman, Wortsman, Ilharco, Gordon, Schuhmann, Schmidt, and Jitsev]{cherti2023reproducible}
Mehdi Cherti, Romain Beaumont, Ross Wightman, Mitchell Wortsman, Gabriel Ilharco, Cade Gordon, Christoph Schuhmann, Ludwig Schmidt, and Jenia Jitsev.
\newblock Reproducible scaling laws for contrastive language-image learning.
\newblock In \emph{IEEE Conf. Comput. Vis. Pattern Recog.}, pages 2818--2829, 2023.

\bibitem[Chiang et~al.(2023)Chiang, Li, Lin, Sheng, Wu, Zhang, Zheng, Zhuang, Zhuang, Gonzalez, Stoica, and Xing]{vicuna2023}
Wei-Lin Chiang, Zhuohan Li, Zi Lin, Ying Sheng, Zhanghao Wu, Hao Zhang, Lianmin Zheng, Siyuan Zhuang, Yonghao Zhuang, Joseph~E. Gonzalez, Ion Stoica, and Eric~P. Xing.
\newblock Vicuna: An open-source chatbot impressing gpt-4 with 90\%* chatgpt quality, 2023.

\bibitem[Chung et~al.(2024)Chung, Hou, Longpre, Zoph, Tay, Fedus, Li, Wang, Dehghani, Brahma, et~al.]{chung2024scaling}
Hyung~Won Chung, Le Hou, Shayne Longpre, Barret Zoph, Yi Tay, William Fedus, Yunxuan Li, Xuezhi Wang, Mostafa Dehghani, Siddhartha Brahma, et~al.
\newblock Scaling instruction-finetuned language models.
\newblock \emph{Journal of Machine Learning Research}, 25\penalty0 (70):\penalty0 1--53, 2024.

\bibitem[Dai et~al.(2024)Dai, Li, Li, Tiong, Zhao, Wang, Li, Fung, and Hoi]{dai2024instructblip}
Wenliang Dai, Junnan Li, Dongxu Li, Anthony Meng~Huat Tiong, Junqi Zhao, Weisheng Wang, Boyang Li, Pascale~N Fung, and Steven Hoi.
\newblock Instructblip: Towards general-purpose vision-language models with instruction tuning.
\newblock \emph{Adv. Neural Inform. Process. Syst.}, 36, 2024.

\bibitem[Devlin et~al.(2018)Devlin, Chang, Lee, and Toutanova]{devlin2018bert}
Jacob Devlin, Ming-Wei Chang, Kenton Lee, and Kristina Toutanova.
\newblock Bert: Pre-training of deep bidirectional transformers for language understanding.
\newblock \emph{arXiv preprint arXiv:1810.04805}, 2018.

\bibitem[Dong et~al.(2024)Dong, Zhang, Zang, Cao, Wang, Ouyang, Wei, Zhang, Duan, Cao, Zhang, Li, Yan, Gao, Zhang, Li, Li, Chen, He, Zhang, Qiao, Lin, and Wang]{internlmxcomposer2}
Xiaoyi Dong, Pan Zhang, Yuhang Zang, Yuhang Cao, Bin Wang, Linke Ouyang, Xilin Wei, Songyang Zhang, Haodong Duan, Maosong Cao, Wenwei Zhang, Yining Li, Hang Yan, Yang Gao, Xinyue Zhang, Wei Li, Jingwen Li, Kai Chen, Conghui He, Xingcheng Zhang, Yu Qiao, Dahua Lin, and Jiaqi Wang.
\newblock Internlm-xcomposer2: Mastering free-form text-image composition and comprehension in vision-language large model.
\newblock \emph{arXiv preprint arXiv:2401.16420}, 2024.

\bibitem[Dosovitskiy et~al.(2020)Dosovitskiy, Beyer, Kolesnikov, Weissenborn, Zhai, Unterthiner, Dehghani, Minderer, Heigold, Gelly, et~al.]{dosovitskiy2020image}
Alexey Dosovitskiy, Lucas Beyer, Alexander Kolesnikov, Dirk Weissenborn, Xiaohua Zhai, Thomas Unterthiner, Mostafa Dehghani, Matthias Minderer, Georg Heigold, Sylvain Gelly, et~al.
\newblock An image is worth 16x16 words: Transformers for image recognition at scale.
\newblock \emph{arXiv preprint arXiv:2010.11929}, 2020.

\bibitem[Gao et~al.(2023)Gao, Zhao, Sun, Xi, Zhang, Ghanem, and Zhang]{gao2023unified}
Qiankun Gao, Chen Zhao, Yifan Sun, Teng Xi, Gang Zhang, Bernard Ghanem, and Jian Zhang.
\newblock A unified continual learning framework with general parameter-efficient tuning.
\newblock In \emph{Int. Conf. Comput. Vis.}, pages 11483--11493, 2023.

\bibitem[Grill et~al.(2020)Grill, Strub, Altch{\'e}, Tallec, Richemond, Buchatskaya, Doersch, Avila~Pires, Guo, Gheshlaghi~Azar, et~al.]{grill2020bootstrap}
Jean-Bastien Grill, Florian Strub, Florent Altch{\'e}, Corentin Tallec, Pierre Richemond, Elena Buchatskaya, Carl Doersch, Bernardo Avila~Pires, Zhaohan Guo, Mohammad Gheshlaghi~Azar, et~al.
\newblock Bootstrap your own latent-a new approach to self-supervised learning.
\newblock \emph{Adv. Neural Inform. Process. Syst.}, 33:\penalty0 21271--21284, 2020.

\bibitem[Guan et~al.(2024)Guan, Liu, Wu, Xian, Li, Liu, Wang, Chen, Huang, Yacoob, Manocha, and Zhou]{guan2023hallusionbench}
Tianrui Guan, Fuxiao Liu, Xiyang Wu, Ruiqi Xian, Zongxia Li, Xiaoyu Liu, Xijun Wang, Lichang Chen, Furong Huang, Yaser Yacoob, Dinesh Manocha, and Tianyi Zhou.
\newblock Hallusionbench: An advanced diagnostic suite for entangled language hallucination and visual illusion in large vision-language models.
\newblock In \emph{IEEE Conf. Comput. Vis. Pattern Recog.}, pages 14375--14385, 2024.

\bibitem[Hadsell et~al.(2006)Hadsell, Chopra, and LeCun]{hadsell2006dimensionality}
Raia Hadsell, Sumit Chopra, and Yann LeCun.
\newblock Dimensionality reduction by learning an invariant mapping.
\newblock In \emph{IEEE Conf. Comput. Vis. Pattern Recog.}, pages 1735--1742. IEEE, 2006.

\bibitem[He et~al.(2020)He, Fan, Wu, Xie, and Girshick]{He_2020_CVPR}
Kaiming He, Haoqi Fan, Yuxin Wu, Saining Xie, and Ross Girshick.
\newblock Momentum contrast for unsupervised visual representation learning.
\newblock In \emph{IEEE Conf. Comput. Vis. Pattern Recog.}, 2020.

\bibitem[Hu et~al.(2022)Hu, Shen, Wallis, Allen-Zhu, Li, Wang, Wang, and Chen]{hu2021lora}
Edward~J Hu, Yelong Shen, Phillip Wallis, Zeyuan Allen-Zhu, Yuanzhi Li, Shean Wang, Lu Wang, and Weizhu Chen.
\newblock Lora: Low-rank adaptation of large language models.
\newblock In \emph{Int. Conf. Learn. Represent.}, 2022.

\bibitem[Kar et~al.(2024)Kar, Tonioni, Poklukar, Kulshrestha, Zamir, and Tombari]{kar2024brave}
O{\u{g}}uzhan~Fatih Kar, Alessio Tonioni, Petra Poklukar, Achin Kulshrestha, Amir Zamir, and Federico Tombari.
\newblock {BRAVE}: Broadening the visual encoding of vision-language models.
\newblock In \emph{Eur. Conf. Comput. Vis.}, 2024.

\bibitem[Li et~al.(2023{\natexlab{a}})Li, Li, Savarese, and Hoi]{blip}
Junnan Li, Dongxu Li, Silvio Savarese, and Steven Hoi.
\newblock Blip-2: Bootstrapping language-image pre-training with frozen image encoders and large language models.
\newblock In \emph{Int. Conf. on Mach. Learning}, pages 19730--19742. PMLR, 2023{\natexlab{a}}.

\bibitem[Li et~al.(2023{\natexlab{b}})Li, Du, Zhou, Wang, Zhao, and Wen]{li2023evaluating}
Yifan Li, Yifan Du, Kun Zhou, Jinpeng Wang, Xin Zhao, and Ji-Rong Wen.
\newblock Evaluating object hallucination in large vision-language models.
\newblock In \emph{Conf. on Empirical Methods in Natural Language Processing}, pages 292--305, Singapore, 2023{\natexlab{b}}. Association for Computational Linguistics.

\bibitem[Lin et~al.(2014)Lin, Maire, Belongie, Hays, Perona, Ramanan, Doll{\'a}r, and Zitnick]{lin2014microsoft}
Tsung-Yi Lin, Michael Maire, Serge Belongie, James Hays, Pietro Perona, Deva Ramanan, Piotr Doll{\'a}r, and C~Lawrence Zitnick.
\newblock Microsoft coco: Common objects in context.
\newblock In \emph{Eur. Conf. Comput. Vis.}, pages 740--755. Springer, 2014.

\bibitem[Liu et~al.(2023{\natexlab{a}})Liu, Li, Li, and Lee]{liu2023improvedllava}
Haotian Liu, Chunyuan Li, Yuheng Li, and Yong~Jae Lee.
\newblock Improved baselines with visual instruction tuning, 2023{\natexlab{a}}.

\bibitem[Liu et~al.(2023{\natexlab{b}})Liu, Li, Wu, and Lee]{liu2023visual}
Haotian Liu, Chunyuan Li, Qingyang Wu, and Yong~Jae Lee.
\newblock Visual instruction tuning.
\newblock In \emph{Adv. Neural Inform. Process. Syst.}, 2023{\natexlab{b}}.

\bibitem[Liu et~al.(2019)Liu, Ott, Goyal, Du, Joshi, Chen, Levy, Lewis, Zettlemoyer, and Stoyanov]{liu2019roberta}
Yinhan Liu, Myle Ott, Naman Goyal, Jingfei Du, Mandar Joshi, Danqi Chen, Omer Levy, Mike Lewis, Luke Zettlemoyer, and Veselin Stoyanov.
\newblock Roberta: A robustly optimized bert pretraining approach.
\newblock \emph{arXiv preprint arXiv:1907.11692}, 2019.

\bibitem[OpenAI(2023)]{openai2023gpt4}
OpenAI.
\newblock Gpt-4 technical report, 2023.

\bibitem[Oquab et~al.(2023)Oquab, Darcet, Moutakanni, Vo, Szafraniec, Khalidov, Fernandez, Haziza, Massa, El-Nouby, et~al.]{oquab2023dinov2}
Maxime Oquab, Timoth{\'e}e Darcet, Th{\'e}o Moutakanni, Huy Vo, Marc Szafraniec, Vasil Khalidov, Pierre Fernandez, Daniel Haziza, Francisco Massa, Alaaeldin El-Nouby, et~al.
\newblock Dinov2: Learning robust visual features without supervision.
\newblock \emph{arXiv preprint arXiv:2304.07193}, 2023.

\bibitem[Peng et~al.(2024)Peng, Wang, Dong, Hao, Huang, Ma, Ye, and Wei]{kosmos2}
Zhiliang Peng, Wenhui Wang, Li Dong, Yaru Hao, Shaohan Huang, Shuming Ma, Qixiang Ye, and Furu Wei.
\newblock Grounding multimodal large language models to the world.
\newblock In \emph{Int. Conf. Learn. Represent.}, 2024.

\bibitem[Radford et~al.(2021)Radford, Kim, Hallacy, Ramesh, Goh, Agarwal, Sastry, Askell, Mishkin, Clark, Krueger, and Sutskever]{pmlr-v139-radford21a}
Alec Radford, Jong~Wook Kim, Chris Hallacy, Aditya Ramesh, Gabriel Goh, Sandhini Agarwal, Girish Sastry, Amanda Askell, Pamela Mishkin, Jack Clark, Gretchen Krueger, and Ilya Sutskever.
\newblock Learning transferable visual models from natural language supervision.
\newblock In \emph{Int. Conf. on Mach. Learning}, pages 8748--8763. PMLR, 2021.

\bibitem[Raffel et~al.(2020)Raffel, Shazeer, Roberts, Lee, Narang, Matena, Zhou, Li, and Liu]{raffel2020exploring}
Colin Raffel, Noam Shazeer, Adam Roberts, Katherine Lee, Sharan Narang, Michael Matena, Yanqi Zhou, Wei Li, and Peter~J Liu.
\newblock Exploring the limits of transfer learning with a unified text-to-text transformer.
\newblock \emph{Journal of machine learning research}, 21\penalty0 (1):\penalty0 5485--5551, 2020.

\bibitem[Research(2024)]{blip3}
Salesforce~AI Research.
\newblock xgen-mm-phi3-mini-instruct model card, 2024.

\bibitem[Schwenk et~al.(2022)Schwenk, Khandelwal, Clark, Marino, and Mottaghi]{aokvqa}
Dustin Schwenk, Apoorv Khandelwal, Christopher Clark, Kenneth Marino, and Roozbeh Mottaghi.
\newblock A-okvqa: A benchmark for visual question answering using world knowledge, 2022.

\bibitem[Sun et~al.(2023)Sun, Fang, Wu, Wang, and Cao]{sun2023evaclipimprovedtrainingtechniques}
Quan Sun, Yuxin Fang, Ledell Wu, Xinlong Wang, and Yue Cao.
\newblock Eva-clip: Improved training techniques for clip at scale, 2023.

\bibitem[Tong et~al.(2024)Tong, Liu, Zhai, Ma, LeCun, and Xie]{Tong_2024_CVPR}
Shengbang Tong, Zhuang Liu, Yuexiang Zhai, Yi Ma, Yann LeCun, and Saining Xie.
\newblock Eyes wide shut? exploring the visual shortcomings of multimodal llms.
\newblock In \emph{IEEE Conf. Comput. Vis. Pattern Recog.}, pages 9568--9578, 2024.

\bibitem[Touvron et~al.(2023)Touvron, Martin, Stone, Albert, Almahairi, Babaei, Bashlykov, Batra, Bhargava, Bhosale, et~al.]{touvron2023llama}
Hugo Touvron, Louis Martin, Kevin Stone, Peter Albert, Amjad Almahairi, Yasmine Babaei, Nikolay Bashlykov, Soumya Batra, Prajjwal Bhargava, Shruti Bhosale, et~al.
\newblock Llama 2: Open foundation and fine-tuned chat models.
\newblock \emph{arXiv preprint arXiv:2307.09288}, 2023.

\bibitem[Villa et~al.(2023)Villa, Alc{\'a}zar, Alfarra, Alhamoud, Hurtado, Heilbron, Soto, and Ghanem]{villa2023pivot}
Andr{\'e}s Villa, Juan~Le{\'o}n Alc{\'a}zar, Motasem Alfarra, Kumail Alhamoud, Julio Hurtado, Fabian~Caba Heilbron, Alvaro Soto, and Bernard Ghanem.
\newblock Pivot: Prompting for video continual learning.
\newblock In \emph{IEEE Conf. Comput. Vis. Pattern Recog.}, pages 24214--24223, 2023.

\bibitem[Villa et~al.(2024)Villa, Alcázar, Soto, and Ghanem]{villa2024magicmerlimmultimodalevaluation}
Andrés Villa, Juan Carlos~León Alcázar, Alvaro Soto, and Bernard Ghanem.
\newblock Behind the magic, merlim: Multi-modal evaluation benchmark for large image-language models, 2024.

\bibitem[Yang et~al.(2024)Yang, Li, Zhou, Song, Wu, Nie, and Ghanem]{yang2024corda}
Yibo Yang, Xiaojie Li, Zhongzhu Zhou, Shuaiwen~Leon Song, Jianlong Wu, Liqiang Nie, and Bernard Ghanem.
\newblock Corda: Context-oriented decomposition adaptation of large language models.
\newblock \emph{arXiv preprint arXiv:2406.05223}, 2024.

\bibitem[Zhai et~al.(2023)Zhai, Mustafa, Kolesnikov, and Beyer]{zhai2023sigmoid}
Xiaohua Zhai, Basil Mustafa, Alexander Kolesnikov, and Lucas Beyer.
\newblock Sigmoid loss for language image pre-training.
\newblock In \emph{Int. Conf. Comput. Vis.}, pages 11975--11986, 2023.

\bibitem[Zhao et~al.(2024)Zhao, Zhang, Chen, Wang, Anandkumar, and Tian]{zhao2024galore}
Jiawei Zhao, Zhenyu Zhang, Beidi Chen, Zhangyang Wang, Anima Anandkumar, and Yuandong Tian.
\newblock Galore: Memory-efficient llm training by gradient low-rank projection, 2024.

\bibitem[Zhou et~al.(2024)Zhou, Cui, Yoon, Zhang, Deng, Finn, Bansal, and Yao]{zhou2023analyzing}
Yiyang Zhou, Chenhang Cui, Jaehong Yoon, Linjun Zhang, Zhun Deng, Chelsea Finn, Mohit Bansal, and Huaxiu Yao.
\newblock Analyzing and mitigating object hallucination in large vision-language models.
\newblock \emph{Int. Conf. Learn. Represent.}, 2024.

\bibitem[Zhu et~al.(2023)Zhu, Chen, Shen, Li, and Elhoseiny]{zhu2023minigpt}
Deyao Zhu, Jun Chen, Xiaoqian Shen, Xiang Li, and Mohamed Elhoseiny.
\newblock Minigpt-4: Enhancing vision-language understanding with advanced large language models.
\newblock \emph{arXiv preprint arXiv:2304.10592}, 2023.

\bibitem[Zou et~al.(2024)Zou, Yang, Zhang, Li, Li, Wang, Wang, Gao, and Lee]{zou2024segment}
Xueyan Zou, Jianwei Yang, Hao Zhang, Feng Li, Linjie Li, Jianfeng Wang, Lijuan Wang, Jianfeng Gao, and Yong~Jae Lee.
\newblock Segment everything everywhere all at once.
\newblock \emph{Adv. Neural Inform. Process. Syst.}, 36, 2024.

\end{thebibliography}
}
\clearpage
\setcounter{page}{1}


\twocolumn[{
\renewcommand\twocolumn[1][]{#1}
\maketitlesupplementary
\thispagestyle{empty}
\begin{center}
  \vspace{-0.6cm}
  \newcommand{\teaserwidth}{\textwidth}
  \centerline{
    \includegraphics[width=\teaserwidth]{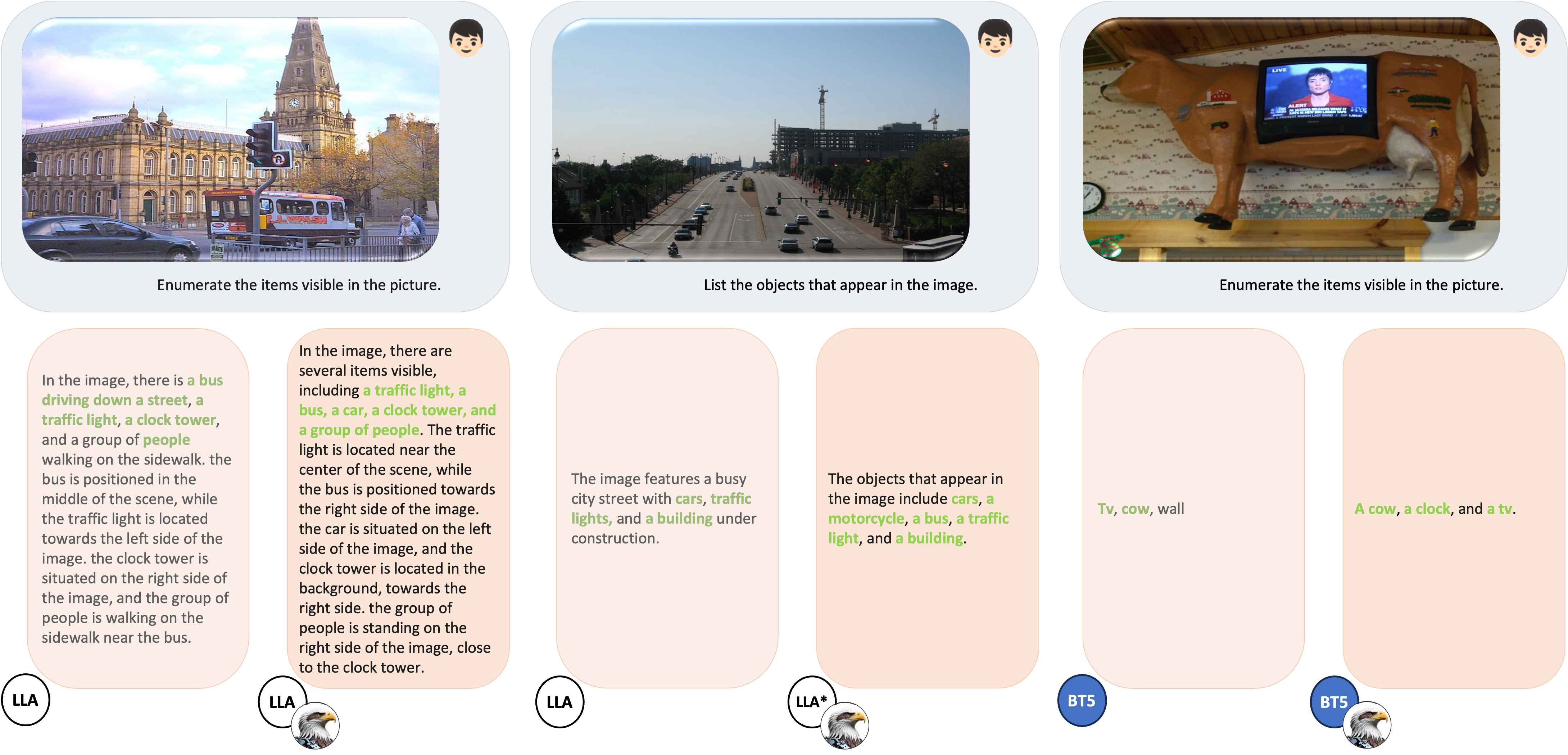}
    \vspace{-0.2cm}
    }
    \captionof{figure}{\textbf{Visual Examples Demonstrating EAGLE's Effectiveness in Reducing Hallucinations in IT-VLMs.} We present three additional illustrative scenarios, each featuring a question about an image processed by a specific IT-VLM using its original visual encoder (left, pink box) and the corresponding EAGLE-tuned visual encoder (right, orange box). The models evaluated include “LLA” (LLaVA-1.5), “LLA*” (LLaVA-1.5*), and “BT5” (BLIP-2 with FlanT5xl). EAGLE demonstrates a significant reduction in hallucinations, providing more visually grounded and reliable descriptions. Correct and incorrect predicted nouns are marked by green and red, respectively.}%
    \vspace{-0.1cm}
  \label{fig:supp_vis_fig}
 \end{center}
}]

\section{Linear Probing of EAGLES-Tuned Models}
\label{sec:lin_prob}

We evaluate whether EAGLE effectively preserves the transfer capability of the features in the original VLM. We compare the zero-shot and linear probing performance of EAGLE-tuned VLMs against the original models. As shown in Table \ref{tab:zst_results_part1}, EAGLE significantly improves the zero-shot accuracy of the feature sequence, with only minor degradation in the zero-shot accuracy of the CLS token. To further validate EAGLE's transfer capability, we analyze the linear probing performance of the visual encoder of both the original and EAGLE-tuned models. For this evaluation, we employ the official implementation provided by \cite{sun2023evaclipimprovedtrainingtechniques}. The analysis includes both CLS token embeddings and sequence embeddings. For sequence embeddings, we compute the logits for each embedding in the sequence and then calculate their average.

As summarized in Table \ref{tab:linear_prob}, EAGLE-tuned models exhibit performance comparable to the original models, even with the CLS token, which remains untrained in our proposed pipeline. This demonstrates EAGLE’s ability to preserve essential global features after training to capture fine-grained object information.

\begin{table}[h]
    \caption{\textbf{Linear Probing performance of EAGLE-tuned VLMs. } We compare the linear probing performance of the visual encoder of both the original and EAGLE-tuned VLMs. EAGLE-tuned models exhibit comparable performance to the original model, considering both the CLS Token and sequence features. }\vspace{-0.2cm}
    
    \resizebox{8.5cm}{!}{
        \begin{tabular}{l ll} 
            \toprule
            \multicolumn{1}{c}{\multirow{2}{*}{\bf Model}} & \multicolumn{2}{c}{\bf IN-1K} \\
            \cmidrule(lr){2-3}
                 & \multicolumn{1}{c}{\bf Cls Acc $\uparrow$} & \multicolumn{1}{c}{\bf Seq. Acc $\uparrow$} \\
            \midrule
            EVA01 ViT-g-14 & \bf 86.50\% & 85.77\% \\
            EAGLE EVA01 ViT-g-14 & 86.07\% \footnotesize\textcolor{red}{(-0.43\%)} & \bf 85.84\% \footnotesize\textcolor{ForestGreen}{(+0.07\%)} \\
            \cmidrule{1-3}
            OpenAI ViT-L-14-336 & \bf 85.15\% & 83.23\% \\
            EAGLE OpenAI ViT-L-14-336 & 83.80\% \footnotesize\textcolor{red}{(-1.35\%)} & \bf 83.51\% \footnotesize\textcolor{ForestGreen}{(+0.28\%)}  \\
            \bottomrule
        \end{tabular}
    }
    \label{tab:linear_prob}
\end{table}

\begin{figure*}[ht]
    \centering
    \includegraphics[width=\textwidth]{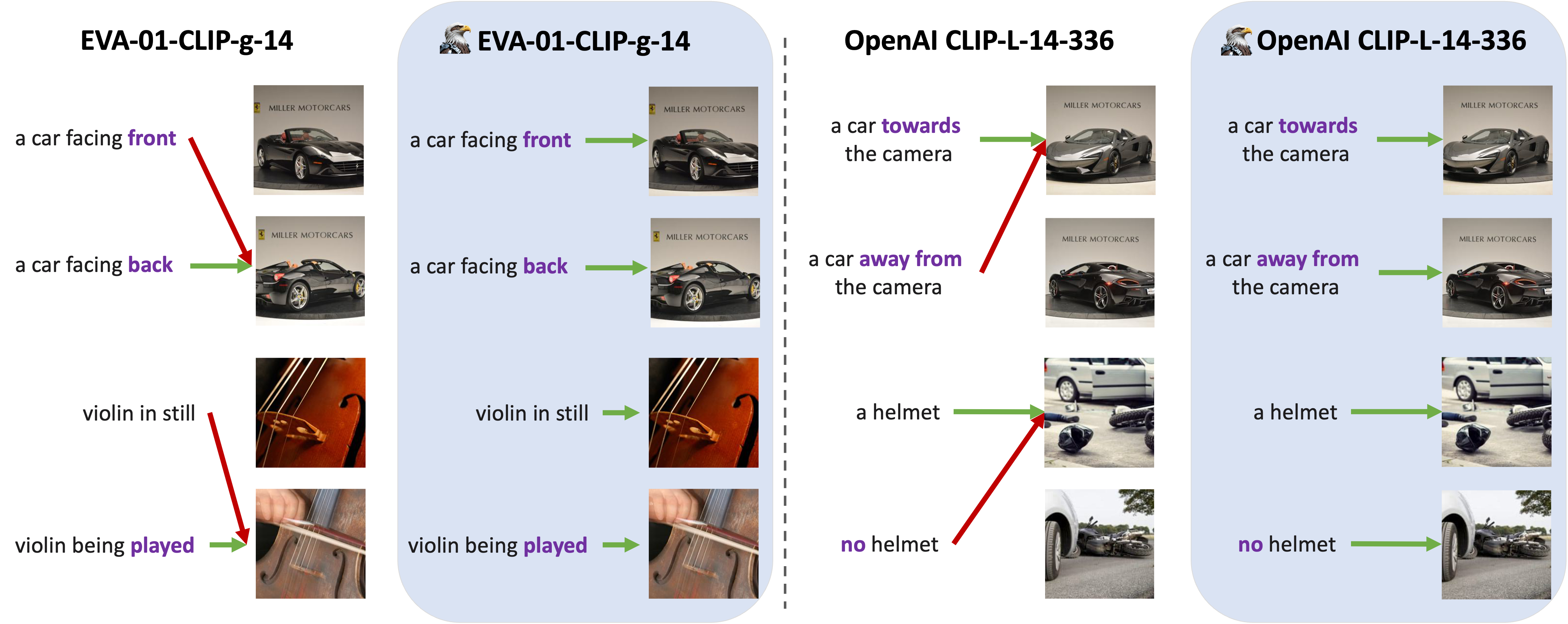}
    \caption{\textbf{Visual Examples of EAGLE Enhancing Visual Grounding of VLMs.} We assess the ability of two VLMs, EVA-01-CLIP-g-14 and OpenAI CLIP-L-14-336, and their corresponding EAGLE-tuned versions (blue boxes) to embed fine-grained visual details in the sequence features, using the MMVP-VLM benchmark. Through two visual examples per model, we show that EAGLE effectively captures subtle visual information in images, enabling it to correctly align image-text pairs even when the images differ only in small, specific features. Correct and incorrect alignments are marked by green and red arrows, respectively.}
    \label{fig:vlm_vis}
\end{figure*}

\section{Qualitative Examples} 
\paragraph{Reducing Hallucinations in IT-VLMs.} Figure \ref{fig:supp_vis_fig} presents 3 additional scenarios when EAGLE effectively reduces the hallucinations of the IT-VLMs. Each scenario features a question about an image by a specific IT-VLM using its original (left pink box) and EAGLE-tuned (right orange box) visual encoder. \textbf{i)} In the first example (left), LLaVA-1.5, equipped with the EAGLE-tuned visual encoder, captures more visual details than its original version, including the car and the spatial arrangement of objects in the image. \textbf{ii)} In the second example (center), our trained LLaVA-1.5 with the EAGLE-tuned visual encoder (LLaVA-1.5*) exhibits improved visual recognition, identifying additional small elements such as the motorcycle and bus in the bottom-left and bottom-right corners of the image, respectively. \textbf{iii)} In the third example (right), EAGLE enhances BLIP-2's visual ability to detect small and partially occluded objects, enabling it to accurately identify the small clock in the bottom-left corner of the image.

\paragraph{More Visual Grounded VLMs.} To further examine the impact of EAGLE on VLMs, we present visual examples from the MMVP-VLM benchmark in Figure \ref{fig:vlm_vis} comparing the ability of the original models and their EAGLE tuned versions to correctly align image-text pairs using the sequence features. As shown in Figure \ref{fig:vlm_vis}, EAGLE-tuned visual encoders can accurately align image-text pairs, even when the images differ in subtle, specific details. These results demonstrate EAGLE’s ability to effectively enhance a VLM’s capacity to embed fine-grained visual information in the sequence features, which is critical for reducing hallucinations in IT-VLMs.

\section{Training Hyperparameters}
Table \ref{tab:hyperparameters} summarizes the additional hyperparameters introduced for EAGLE training, building on the original settings provided by \cite{sun2023evaclipimprovedtrainingtechniques} for training EVA01-CLIP-g-14. The same hyperparameters are applied to both VLMs, EVA01-CLIP-g-14 and OpenAI CLIP-L-14-336. As detailed in Section \ref{sec:exp_sec}, we tune these models using two A100 GPUs (80GB).

\begin{table}[h]
    \centering
    \caption{\textbf{EAGLE Hyperparameters. } We provide the hyperparameters we add to the original setting provided by \cite{sun2023evaclipimprovedtrainingtechniques} for EAGLE training.}\vspace{-0.2cm}
    
    \resizebox{5cm}{!}{
        \begin{tabular}{l l} 
            \toprule
            \multicolumn{1}{c}{\bf Hyperparameter} &  \\
            \midrule
            Lr & 4e-6 \\
            Batch size & 512 \\
            Warmup & 25000 \\
            Optimizer & GaLoreAdamW \\
            GaLore Rank & 128 \\
            GaLore scale & 0.25 \\
            GaLore Projection Type & std \\
            seed & 4096 \\
            \bottomrule
        \end{tabular}
    }
    \label{tab:hyperparameters}
\end{table}

\end{document}